\documentclass[letterpaper, 10 pt, conference]{ieeeconf}  

\IEEEoverridecommandlockouts                              

\usepackage{todonotes}
\usepackage{support-caption}
\usepackage{subcaption}
\usepackage{amsmath}
\usepackage{comment}
\usepackage{multirow} 

\usepackage[normalem]{ulem}

\usepackage{titlesec}
\titleformat{\paragraph}[runin]
{\normalfont\normalsize\bfseries}{}{1em}{}
\titlespacing{\paragraph}
{0pt}{3.25ex plus 1ex minus .2ex}{1em}

\usepackage{scalerel}
\usepackage{tikz}
\usetikzlibrary{svg.path}

\definecolor{orcidlogocol}{HTML}{A6CE39}
\tikzset{
  orcidlogo/.pic={
    \fill[orcidlogocol] svg{M256,128c0,70.7-57.3,128-128,128C57.3,256,0,198.7,0,128C0,57.3,57.3,0,128,0C198.7,0,256,57.3,256,128z};
    \fill[white] svg{M86.3,186.2H70.9V79.1h15.4v48.4V186.2z}
                 svg{M108.9,79.1h41.6c39.6,0,57,28.3,57,53.6c0,27.5-21.5,53.6-56.8,53.6h-41.8V79.1z M124.3,172.4h24.5c34.9,0,42.9-26.5,42.9-39.7c0-21.5-13.7-39.7-43.7-39.7h-23.7V172.4z}
                 svg{M88.7,56.8c0,5.5-4.5,10.1-10.1,10.1c-5.6,0-10.1-4.6-10.1-10.1c0-5.6,4.5-10.1,10.1-10.1C84.2,46.7,88.7,51.3,88.7,56.8z};
  }
}

\newcommand\orcidicon[1]{\href{https://orcid.org/#1}{\mbox{\scalerel*{
\begin{tikzpicture}[yscale=-1,transform shape]
\pic{orcidlogo};
\end{tikzpicture}
}{|}}}}

\usepackage{hyperref}
\usepackage{float}

\title{\huge ClearLines - Camera Calibration from Straight Lines}

\author{Gregory Schroeder$^{1, 2}$\orcidicon{0009-0005-7340-1715} \textit{Member, IEEE}, Mohamed Sabry$^{1}$\orcidicon{0000-0002-9721-6291} \textit{Member, IEEE}, \\ and Cristina Olaverri-Monreal$^{1}$\orcidicon{0000-0002-5211-3598} \textit{Senior Member, IEEE}%
\thanks{$^1$ Johannes Kepler University Linz, Austria, Department Intelligent
Transport Systems, Altenberger Straße 69, 4040 Linz, Austria.
\texttt{\{gregory.schroeder, mohamed.sabry, cristina.olaverri-monreal\}@jku.at}}%
\thanks{$^{2}$Intelligent Systems Functions Department, IAV GmbH, Berlin, Germany 
{\tt\footnotesize gregory.schroeder@iav.de}}%
}

\begin{document}
\maketitle
\thispagestyle{empty}
\pagestyle{empty}

\setlength{\abovecaptionskip}{4pt}   

\def\datasetname{ClearLines} 
\def\urldataset{https://github.com/gregory-schroeder/clearlines_dataset.git}

\begin{abstract}
The problem of calibration from straight lines is fundamental in geometric computer vision, with well-established theoretical foundations. However, its practical applicability remains limited, particularly in real-world outdoor scenarios. These environments pose significant challenges due to diverse and cluttered scenes, interrupted reprojections of straight 3D lines, and varying lighting conditions, making the task notoriously difficult. Furthermore, the field lacks a dedicated dataset encouraging the development of respective detection algorithms. In this study, we present a small dataset named "\datasetname", and by detailing its creation process, provide practical insights that can serve as a guide for developing and refining straight 3D line detection algorithms.
\end{abstract}


\section{Introduction}
\label{sec:introduction}

At the time of writing, the KITTI dataset~\cite{2013-kitti-dataset} was published 12 years ago. For the past three years, the best performing algorithm in its odometry/slam challenge has been the camera-based work by Cvišić et al.~\cite{kittical}. Notably, they did not develop a new slam algorithm in this work but focused on finding a better set of camera calibration parameters, which highlights the importance of intrinsic camera calibration for visual SLAM and odometry algorithms.
%
%
%
%
%
While an exceptionally well summarized theoretical framework for geometric computer vision exists (see e.g.~\cite{hartley2003multiple, 2011-overview-sturm}), it took nearly a decade to identify improved camera parameters for one of the most renowned datasets~\cite{2013-kitti-dataset}. 
Interestingly, this refinement~\cite{kittical} did not directly leverage the established theoretical framework. Instead, the dataset's calibration images, IMU data and image sequences together with a comprehensive SLAM pipeline, were employed in a labor and computationally expensive process tailored to the dataset;
raising questions about the practical applicability and limitations of online camera calibration theories.

Our work addresses one of the areas where this theoretical-practical gap is particularly evident: calibration using straight lines. The underlying theory is simple — straight lines in 3D reproject to straight lines in 2D under the pinhole camera model.
Deviations from this assumption reveal the non-perspective components of the camera model, namely the distortion parameters, which can then be estimated.
%
%
The idea of utilizing 3D lines for calibration was already introduced more than half a century ago in~\cite{1971-64-brown-first-plumb-line}.
Since then, numerous improvements regarding accuracy, computational efficiency, finding closed-form solutions or adaptions to different camera- and distortion-models have been proposed
~\cite{2010-241-equidistant-fish-eye, 2009-520-simple-method-radial-distortion, 2008-521-new-calibration-model}.
%
These works demonstrate excellent calibration results, but need to use synthetic images or images of calibration patterns, require human supervision or manual selection of line points.
Other works also take overall robustness into account, which enables the usage of real imagery~\cite{2003-127-Nonmetric-lens-distortion, 2003-503-robust-calibration, 2009-520-simple-method-radial-distortion}.
%
Yet, the imagery still needs to display beneficial indoor environments with significant, continuous edges and otherwise mostly unstructured areas to limit the introduction of outliers 
The application of 3D-line based calibration is hindered in areas where taking controlled, indoor images is not feasible. 
For real outdoor applications, the task faces multiple difficulties, including variation in illumination, diverse scene content, as well as cluttered and non-continuous re-projection of straight 3D lines. 
These challenges are particularly relevant in the fields of Intelligent Transportation Systems, Autonomous Vehicles, and Platooning~\cite{zhang2024robust}, and must be addressed in automated software stacks~\cite{sabry2024autonav}.
Accurate calibration is essential for downstream tasks such as visual SLAM, localization, and map alignment, where distortion errors can accumulate and degrade system performance.

Furthermore, an application such as edge-segment detection, as defined below, is a multi-step process, where the strategy of selecting the best-performing algorithm for each stage is ineffective. The objectives of individual tasks can diverge drastically from the objectives of edge-segments detection. 
Consequently, results considered successful in their domain, may fail to contribute to the overall goal. 
%
%
%
For example, edge detection, which typically serves as the initial stage in edge-segment pipelines, focuses on achieving a comprehensive set of precise edge pixels. While effective for general tasks such as image segmentation or object proposal generation, this objective differs from the goal of edge-segment detection in the context of camera calibration: identifying long, smooth, and continuous edge-segments that correspond to 3D lines.
%
Despite recent advances in edge detection and the availability of benchmark datasets for training and evaluating edge detection algorithms, no dedicated dataset exists for evaluating edge-segment detection in the context of camera calibration. 
This lack of datasets stems from the impracticality of manual labeling, as it requires accurately labeling hundreds of pixels to represent a single edge-segment. This high level of precision and effort makes creating ground truth references for straight edge-segments exceedingly difficult.
To address this gap, we present "\datasetname", a small dataset designed specifically for edge-segment detection and provide practical guidance for implementing pipelines tailored to this purpose.
%
%

\vspace{-10pt}
\paragraph{Definitions and Notation:}
\begin{itemize}
    \item \textbf{Edge:}
An edge is a binary representation in an image where pixel values of 1 indicate the presence of edges, typically identified by intensity gradients. Although useful for general image analysis, raw edge images lack the structural organization required for higher-level tasks such as camera calibration.

    \item \textbf{Edge-Segment:}  
Edge-segments are refined representations of image edges, formed by grouping edge pixels into coherent structures. These structures are obtained by applying edge-detection algorithms and subsequent post-processing steps, such as edge-linking or edge-chaining. Edge-segments serve as the basis for fitting geometric primitives like lines, circles, and ellipses. However, when derived from outdoor scenes, most edge-segments do not correspond to straight lines in 3D, rendering them unsuitable for camera calibration.

    \item \textbf{Straight Edge-Segment:}  
A straight edge-segment is a specific type of edge-segment that corresponds to a straight line in 3D space. These segments are geometrically meaningful and can be used for camera calibration. 
In general, their projections in the image space do not appear as straight lines due to camera distortions.
For brevity, we will refer to them simply as "straight edge-segments" throughout the paper.
\end{itemize}

\vspace{-10pt}

\paragraph{Contributions:}
\begin{itemize}
    \item We introduce the compact "\datasetname" dataset for 3D lines in typical outdoor scenes, sampled from the KITTI dataset~\cite{2013-kitti-dataset} and the IAMCV dataset~\cite{2023-JKU-dataset, 2024-JKU-dataset-2} aimed for the evaluation of camera calibration quality.
    \item We provide user-friendly evaluation scripts designed to measure performance metrics and validate results derived from this dataset.
    \footnote{For more details, see the repository at 
    {\scriptsize\url{https://gitlab.com/intelligent-transportation-systems/pdrive/clearlines_dataset}}}
    \item We offer practical guidance for implementing a pipeline to detect straight edge-segments effectively.
\end{itemize}

\vspace{-10pt}

\paragraph{Paper Outline:}
%
Section~\ref{sec:related_work} surveys existing literature. 
%
Section~\ref{sec:dataset_creation} details the design and setup of the dataset, including metrics and challenges encountered.
%
Section~\ref{sec:methodology} provides an overview of our detection pipeline employed as a prelabeling step. It offers practical recommendations and best practices. 
%
%
Section~\ref{sec:discussion} evaluates the utility and limitations of the proposed dataset. We address areas where the dataset could be improved or extended to better serve the research community.
%
Finally, Section~\ref{sec:conclusion} highlights key contributions and conclusions.


\section{Related Work} 
\label{sec:related_work}


%
%


%

Bogdan et al.~\cite{DeepCalib-2018} utilize publicly available panorama images from the Internet to generate synthetic images through a virtual camera. These synthetic images form a dataset used to train a neural network. The method is effective for applications where approximate undistortion is sufficient, as it employs only a single distortion parameter.
%
%
With regard to online camera calibration datasets, previous research focused mainly on sports related applications \cite{magera2024universal}, such as predicting the camera parameters using the re-projection error based on the landmarks of a football court \cite{theiner2023tvcalib}. Other datasets focused on Basketball \cite{sha2020end} and hockey related data \cite{jiang2020optimizing}. These approaches take into account the structure of the courts, their definitive landmarks, as well as the camera locations to help with the camera calibration tasks. However, these approaches utilize clear, fixed structures that are barely occluded. For automated driving applications, these approaches cannot be utilized. 

Automated data generation for line detection, as proposed by \cite{2021-Learnable-Line-Detector}, leverages fully calibrated and synchronized camera-LiDAR setups with precise ground truth poses. 
Assuming that the camera distortion model is accurate, such methods could serve as a foundation for straight edge-segment datasets; which is conceptually similar to our detection pipeline presented in Section~\ref{sec:methodology}, which is employed for pre-labeling.


Recent advances in edge detection have focused mainly on deep learning-based methods
, achieving results that rival or surpass human capabilities. For a comprehensive overview of these developments, see~\cite{2022-survey-edge-detect-learning-base, 2022-survey-edge-detect}.
The performance of edge detection algorithms is typically evaluated using three main datasets: Multicue~\cite{Multicue-edge-detect-dataset}, NYUD~\cite{NYUD-edge-detect-dataset}, and the widely-used BSDS500 dataset~\cite{BSDS500}. 
These datasets provide benchmarks for training and assessing the detection of general edges, with results ranked by their F-score—the harmonic mean of precision and recall. This metric is widely accepted as it aligns well with the requirements of related tasks, such as image segmentation and object proposal generation.
%
%
%
To the knowledge of the authors, no datasets exist for straight-line camera calibration, particularly in the context of automated driving. 
%

\section{ClearLines Dataset} 
\label{sec:dataset_creation}



\begin{figure*}
\def\heightfigJKU{3.3cm}
\subfloat{\includegraphics[height=\heightfigJKU,width=5.85cm]{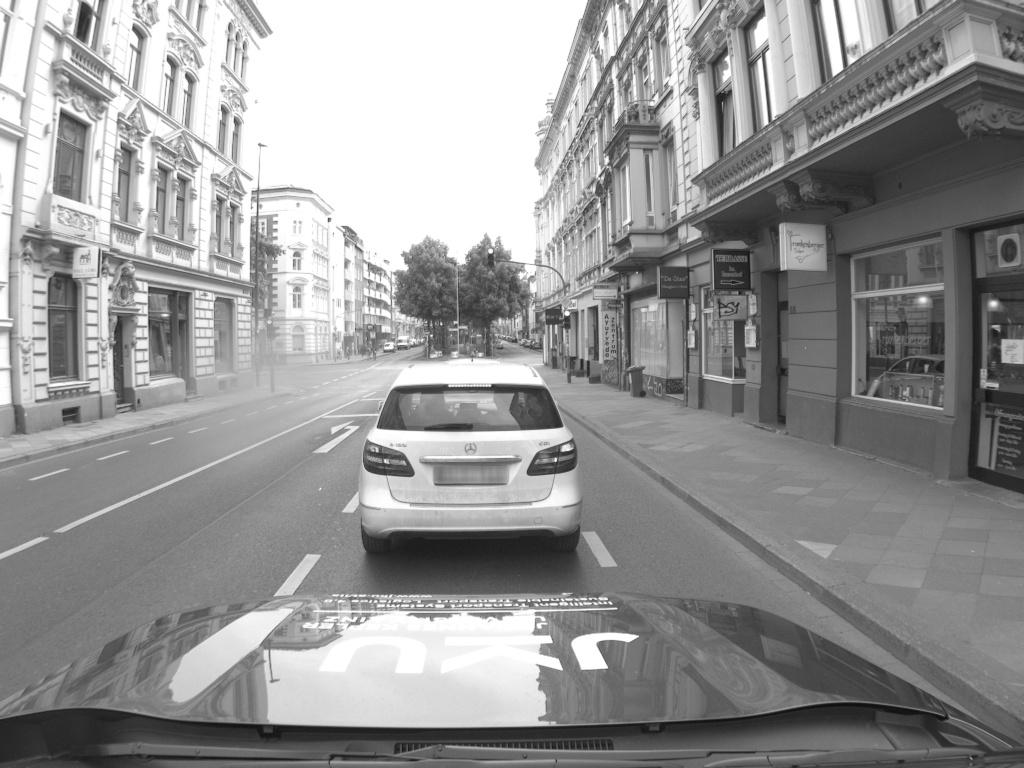}}
\hspace{0.01cm}
\subfloat{\includegraphics[height=\heightfigJKU,width=5.85cm]{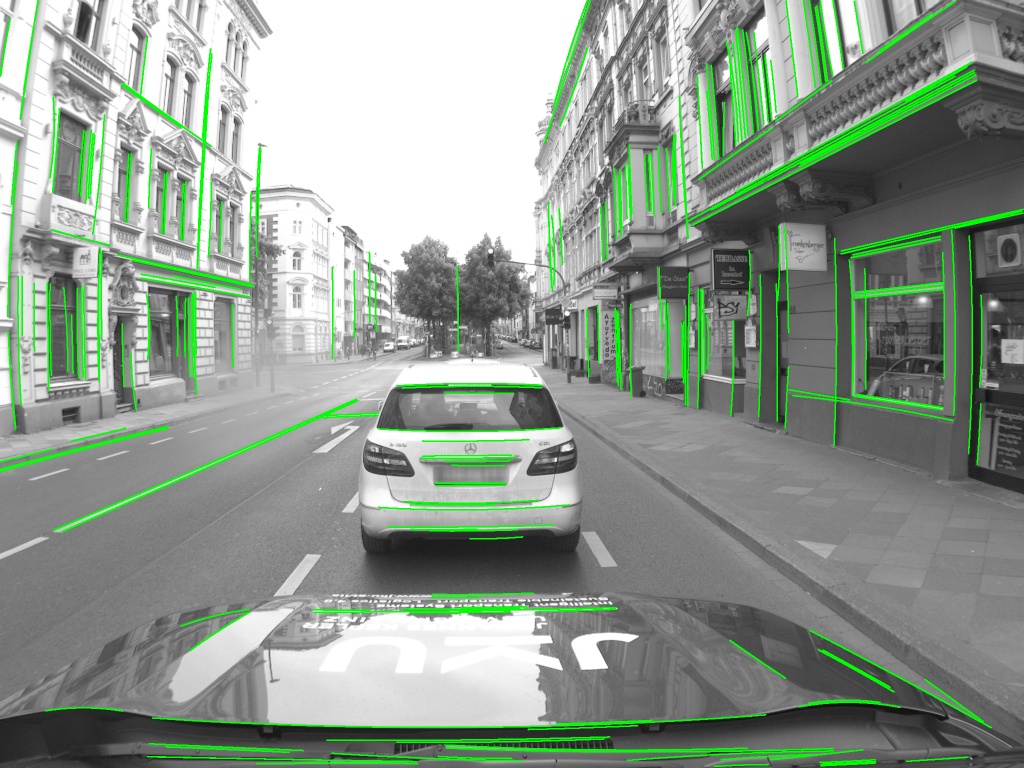}}
\hspace{0.01cm}
\subfloat{\includegraphics[height=\heightfigJKU,width=5.85cm]{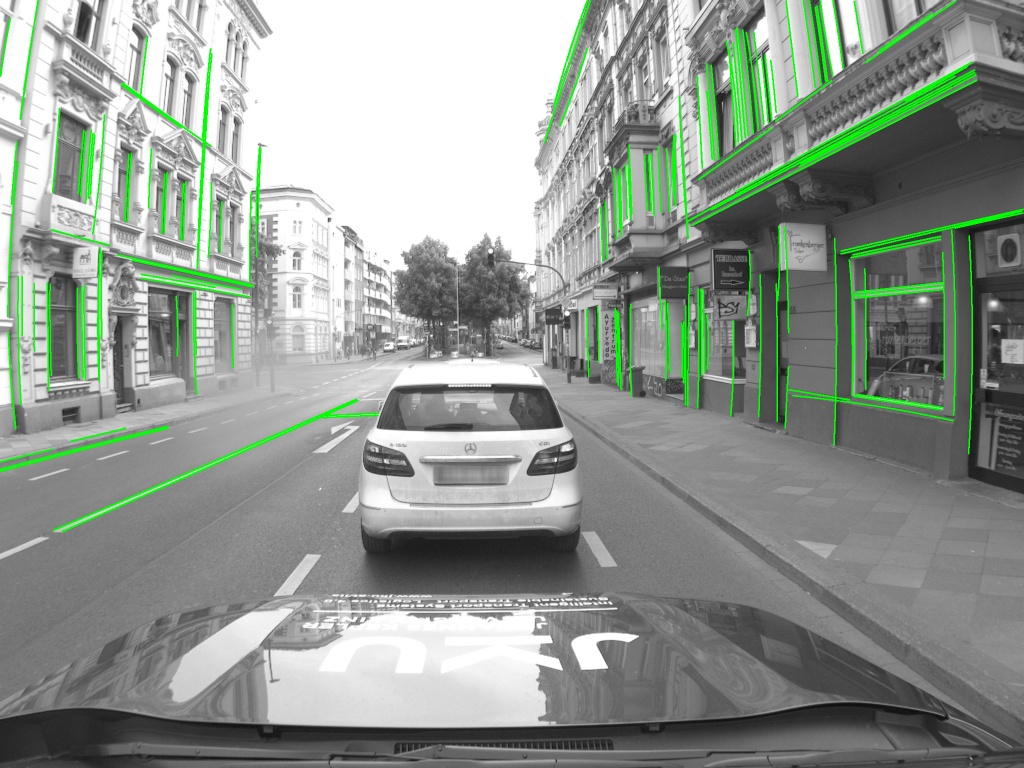}}
\vspace{0.10cm}
\subfloat{\includegraphics[height=\heightfigJKU,width=5.85cm]{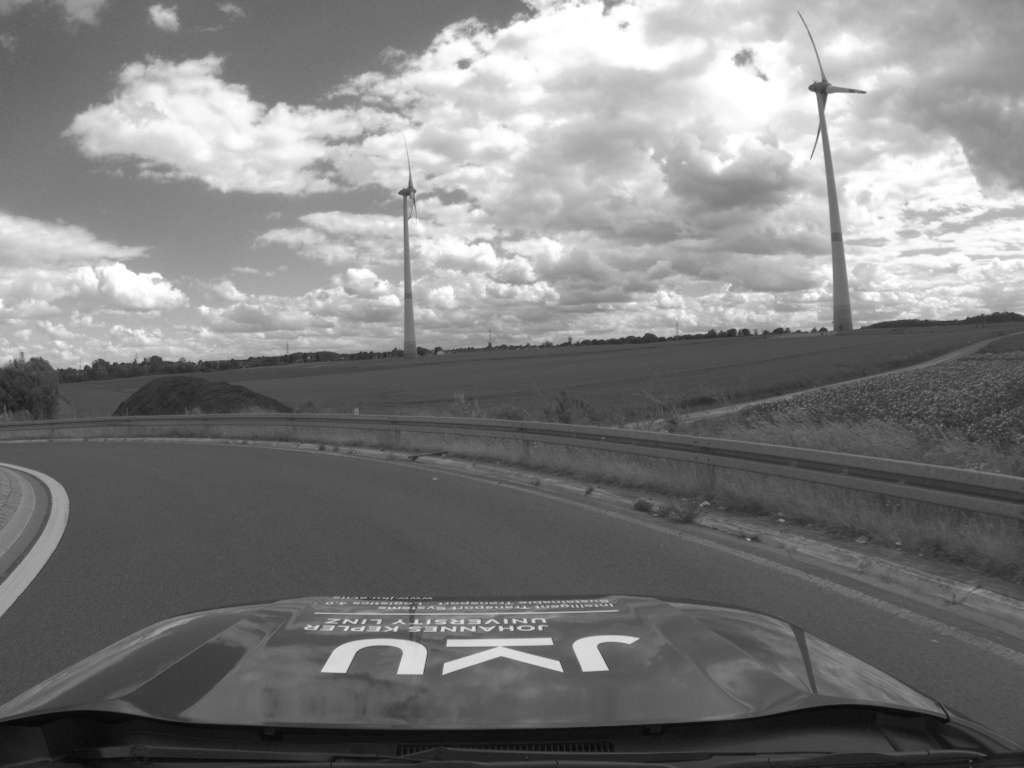}}
\hspace{0.01cm}
\subfloat{\includegraphics[height=\heightfigJKU,width=5.85cm]{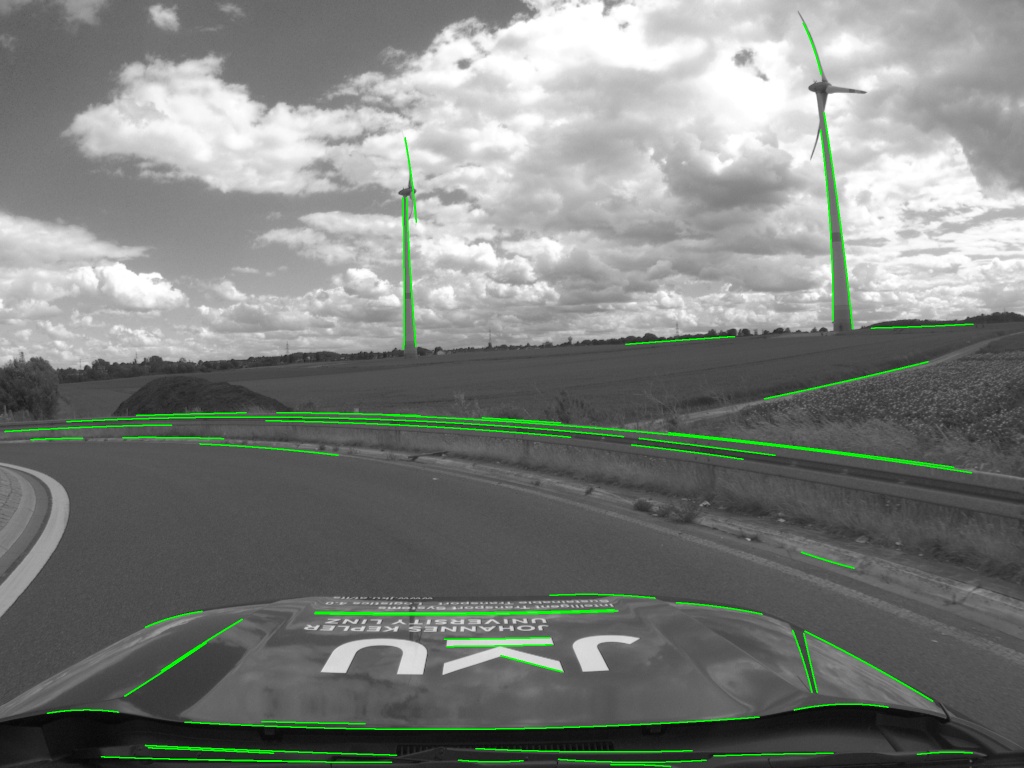}}
\hspace{0.01cm}
\subfloat{\includegraphics[height=\heightfigJKU,width=5.85cm]{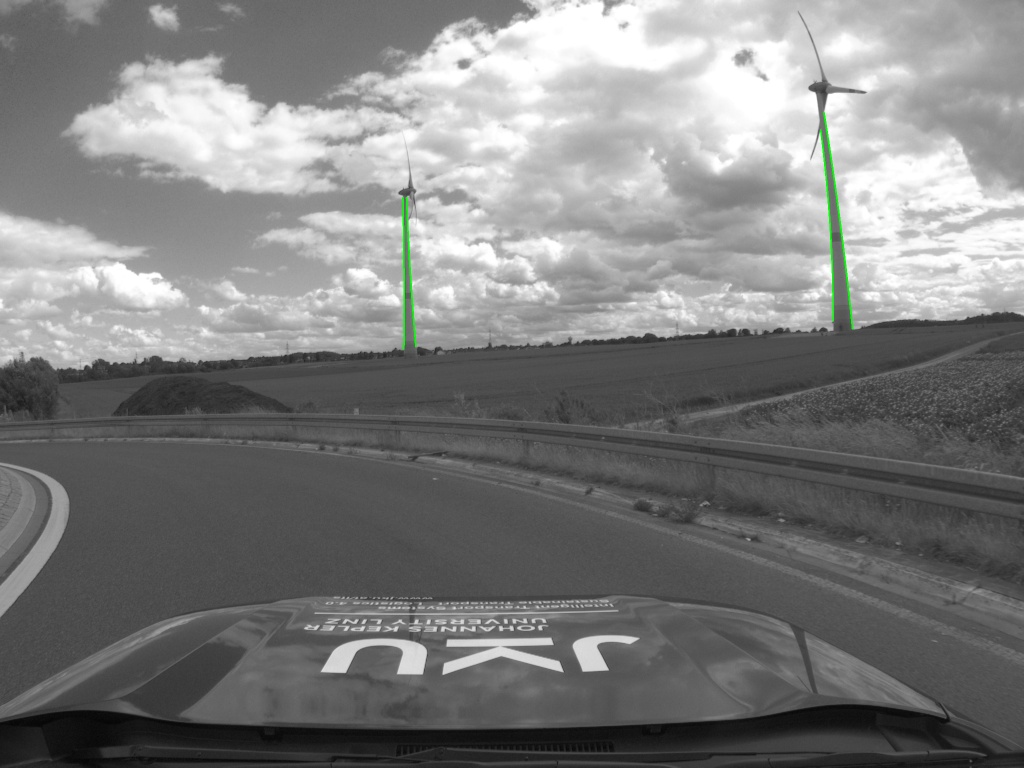}}
\vspace{0.10cm}
%
\subfloat{\includegraphics[height=\heightfigJKU,width=5.85cm]{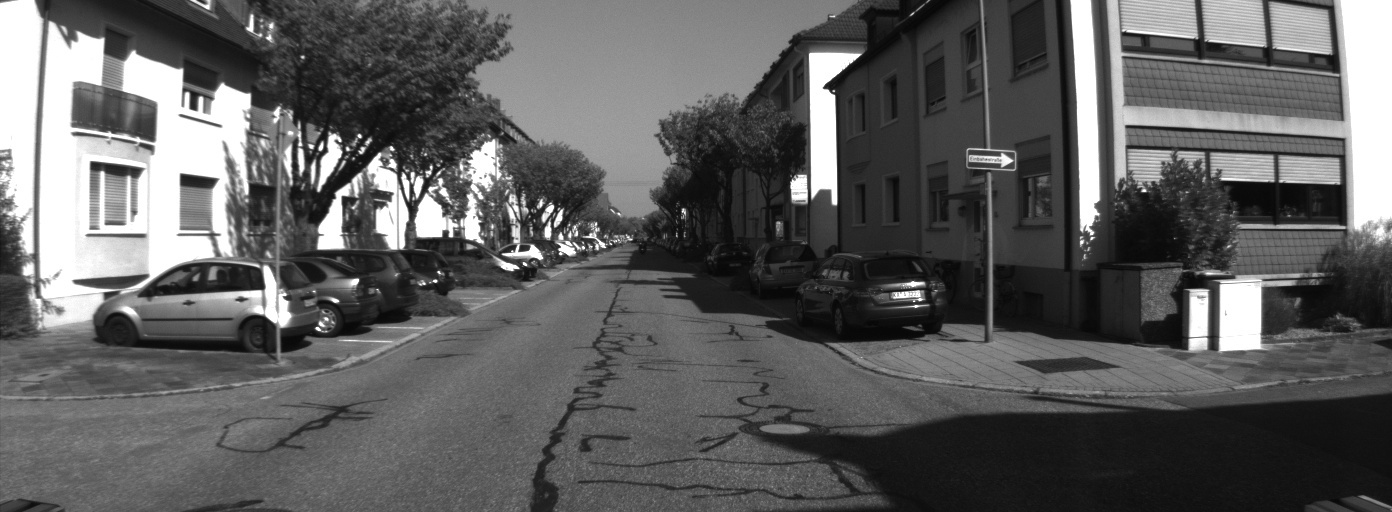}}
\hspace{0.01cm}
\subfloat{\includegraphics[height=\heightfigJKU,width=5.85cm]{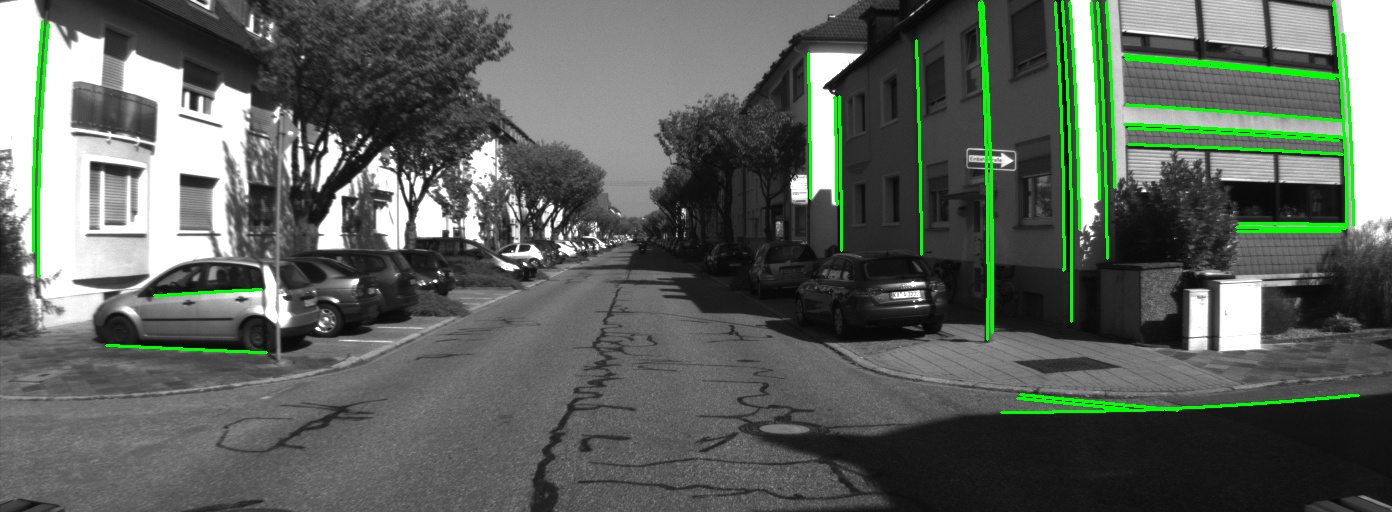}}
\hspace{0.01cm}
\subfloat{\includegraphics[height=\heightfigJKU,width=5.85cm]{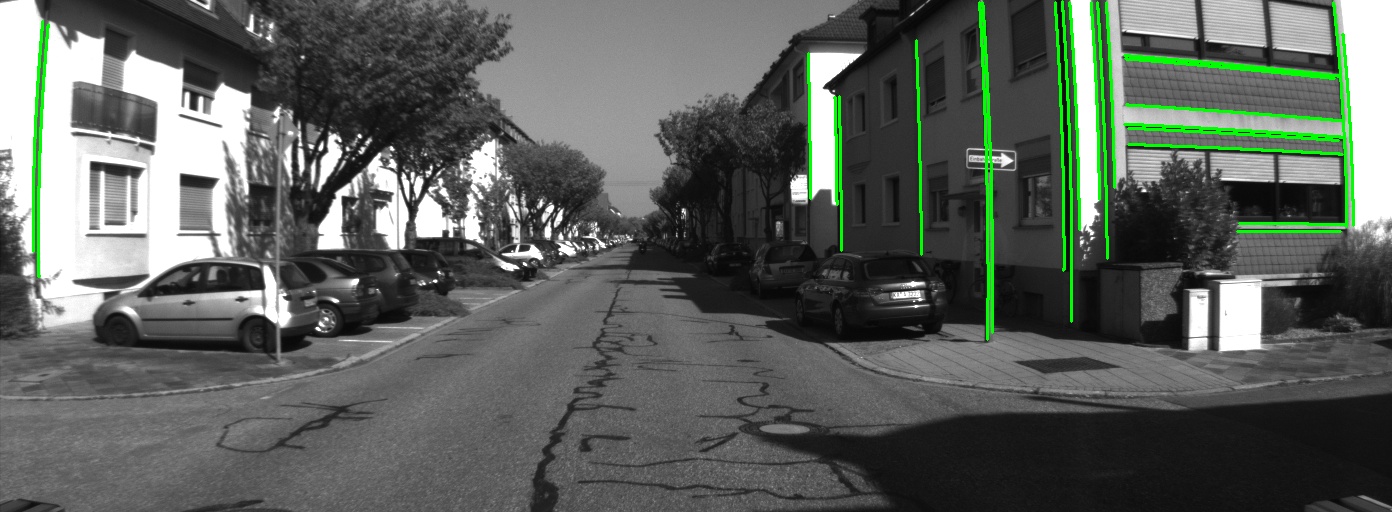}}
%
%
%
%
\caption{Examples from the "\datasetname" dataset. First column: Original images (from IAMCV dataset~\cite{2023-JKU-dataset} and KITTI dataset~\cite{2013-kitti-dataset}). Second column: output of our edge-segment detection pipeline. Third column: Results after manual filtering. All images in black-white for visualization purpose}
\label{fig:dataset_examples}
\end{figure*}

In order to create the dataset, images were taken from multiple datasets containing the vehicle view perspective, as well as being used in various automated driving applications, such as visual odometry. KITTI~\cite{2013-kitti-dataset} was selected as the first source of data as it is well established in the field with data that can be utilized in various automated driving modules. The data was collected in Karlsruhe, Germany, mostly in an urban environment.
For the second source of data, the IAMCV dataset~\cite{2023-JKU-dataset, 2024-JKU-dataset-2} was used. This dataset was collected using the Johannes Kepler University, Intelligent Transport Systems research vehicle in real-world driving scenarios that include roundabouts, intersections, country roads, and highways, recorded across diverse locations in Germany.
At the time of writing 85 images from KITTI, 50 images from IAMCV downtown and 21 images from IAMCV rural/ roundabout were selected and labeled (see Table~\ref{tab:dataset_summary} for a summary).

\begin{table}
\centering
\caption{Characteristics of the ClearLines Dataset}
\begin{tabular}{|p{2.5cm}|p{3.5cm}|p{1.5cm}|}
\hline
\textbf{Dataset Source} & \textbf{Environment Type}                      & \textbf{Number of Images} \\ \hline
KITTI                  & Urban (Karlsruhe, Germany)                    & 85                        \\ \hline
IAMCV (Downtown)       & Urban environments (various locations, Germany) & 50                        \\ \hline
IAMCV (Rural/Roundabout) & Rural roads, roundabouts, highways            & 21                        \\ \hline
\end{tabular}
\label{tab:dataset_summary}
\end{table}


The process of manually labeling all straight edge-segments in a single image is a challenging, time-consuming task. This is apparent, as a single, useful edge-segment consists of hundreds of sub-pixel accurate image positions that need to be accurately labeled. Therefore, the selected data frames from the aforementioned datasets, were processed with an edge-segment detection pipeline (see section~\ref{sec:methodology}), which works well in arbitrary outdoor scenes and under significant camera distortion. The pipeline is tuned to have a high recall, which comes with an inevitable loss in precision. 
%
%
The images that did not retain a recall score close to a 100\,\% (assessed through visual inspection) were discarded. 
This process serves as a pre-labeling step. The minimum length of a usable edge-segment is set to 100 pixels, based on empirical observations of calibration reliability. 
The output of this pipeline is manually filtered to only retain edge-segments corresponding to straight 3D-lines. 
%
Figure~\ref{fig:dataset_examples} shows examples of our dataset and also illustrates common false positives which need to be manually removed. 
Without context it is nearly impossible to distinguish edge-segments on cars and lane markings, which are close to straight in 3D, from true straight edge-segments.


\vspace{5pt}
\textbf{Metrics:}
As discussed in Section~\ref{sec:introduction}, numerous studies have demonstrated that effective camera calibration can be achieved once a sufficiently large set of straight edge-segments is detected. The precise localization of edge-segments in the image is not a limiting factor, and we summarize a sub-pixel accurate approach in Section~\ref{sec:methodology}. Instead, two main challenges arise when detecting straight edge-segments in real-world outdoor data:

i) False Positives: Many detected edge-segments do not correspond to actual straight edges in the scene. Traditional approaches, as discussed in Section~\ref{sec:introduction}, struggle with high false positive rates that cannot be effectively handled.

ii) Incomplete Straight Edge-Segments: Due to factors such as occlusions by small foreground objects or lighting-induced discontinuities, straight edges often appear fragmented in the image. These incomplete segments reduce the signal-to-noise ratio, particularly for shorter segments, which are less effective for calibration.
\newline
%
To evaluate the performance of straight edge-segment detection under these conditions, we adopt the precision-recall framework, which is commonly applied in object detection tasks. These metrics are well-suited to addressing the aforementioned challenges:
%

\begin{itemize}\setlength{\itemsep}{0pt}
    \item \textbf{Precision:} Measures the proportion of detected edge-segments that are true straight edge-segments. It is calculated as \( \text{Precision} = \frac{\text{True Positives}}{\text{True Positives} + \text{False Positives}} \).
    \vspace{25pt}
    \item \textbf{Recall:} Measures the proportion of all true straight edge-segments that are successfully detected. It is calculated as \( \text{Recall} = \frac{\text{True Positives}}{\text{True Positives} + \text{False Negatives}} \).
\end{itemize}






\vspace{-10pt}

\begin{figure}
 \includegraphics[height=4cm, width=0.99\linewidth]{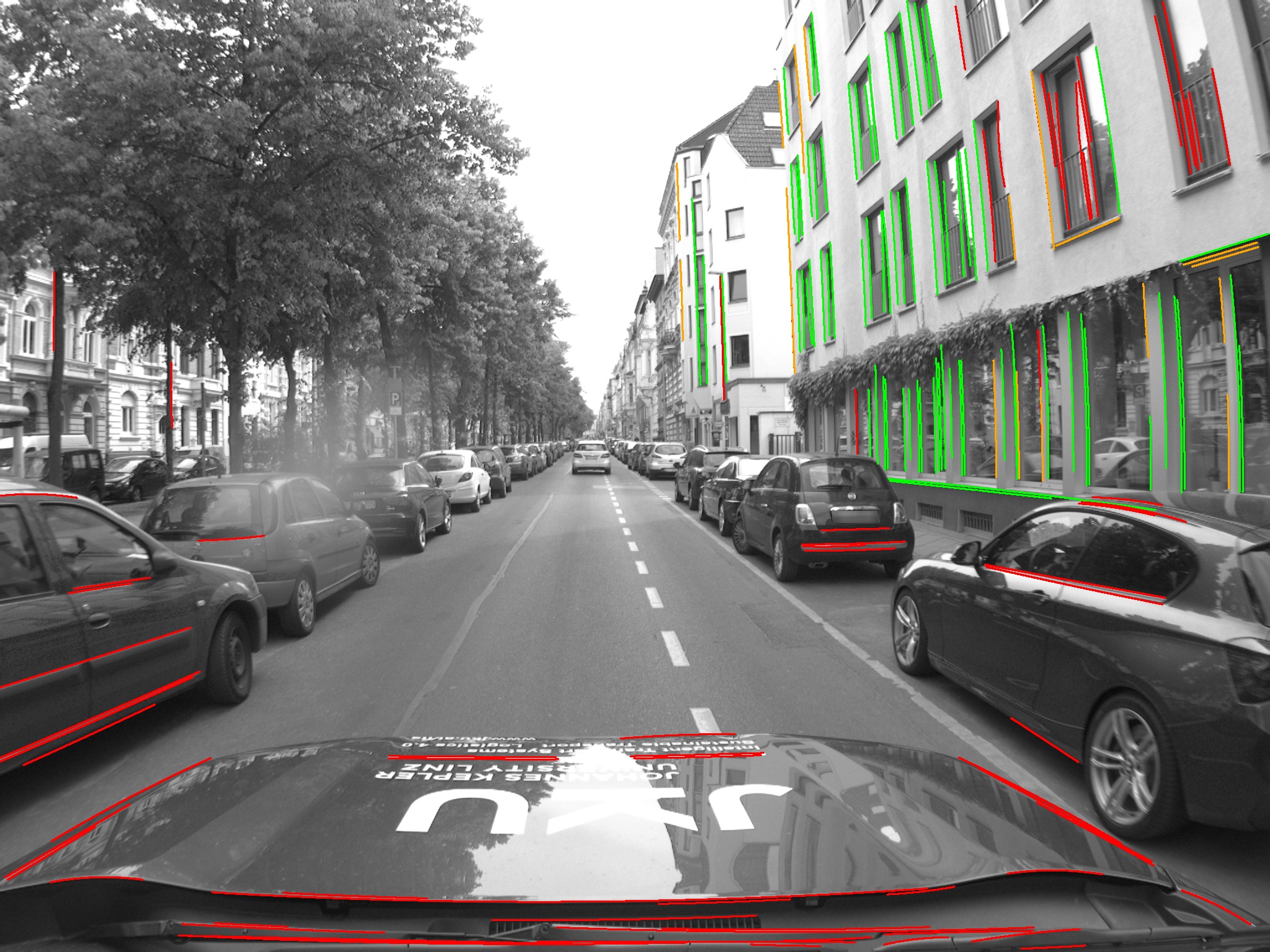}
\caption{Example output from the "\datasetname" evaluation framework. Green: True positives, Red: False positives, Orange: False negatives. Source: IAMCV dataset~\cite{2024-JKU-dataset-2}.} \label{fig:example_evaluation}
\end{figure}

\paragraph{Evaluation:}
To evaluate detection performance against our dataset, users can utilize the evaluation script provided in the repository (see README for details
\footnote{\scriptsize\url{https://gitlab.com/intelligent-transportation-systems/pdrive/clearlines_dataset}}
Besides calculating the metrics mentioned above, the framework visualizes the performance of the detection algorithm by highlighting true positives, false positives, and missed detections. This visualization provides a comprehensive analysis of the algorithm’s strengths and weaknesses (see Figure~\ref{fig:example_evaluation} for an example).

\section{Pipeline Overview} 
\label{sec:methodology}


\begin{figure*}
\includegraphics[height=2.4cm,width=17.5cm]{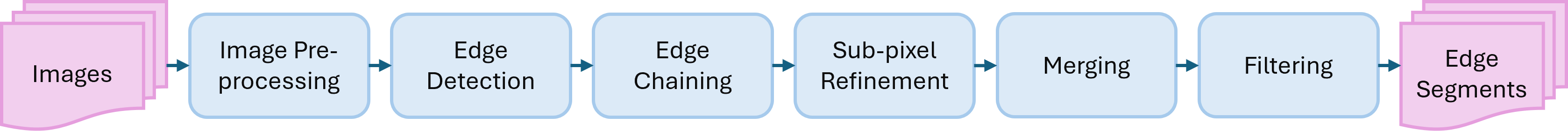}
\caption{High-level overview of the edge-segment detection pipeline. Pink indicates input and output data, while blue represents processing steps. A detailed description is provided in the text.}
\label{fig:flowchart_pipeline}
\end{figure*}

\begin{table}[h]
\caption{\small Precision and recall achieved by our pre-labeling pipeline. The pipeline was tuned for high recall, and only images with near-perfect recall were retained, as removing false positives is more practical than adding missed detections.}\label{tab:metrics}
\centering
\begin{tabular}{lcc}
\hline
\textbf{Dataset} & \textbf{Precision} & \textbf{Recall} \\
\hline
\datasetname \, - KITTI & 0.43 & 0.91 \\
\datasetname \, - IAMCV downtown & 0.46 & 0.98 \\
\datasetname \, - IAMCV roundabout & 0.24 & 0.98 \\
\hline
\end{tabular}
\end{table}

This section describes the methodology of our straight edge-segment detection pipeline.
Table~\ref{tab:metrics} summarizes the precision and recall achieved by this pipeline.
As highlighted in Section~\ref{sec:related_work}, the absence of datasets for validating edge segment detection in camera calibration, coupled with the ineffectiveness of selecting domain-best algorithms for each stage, complicates the justification of methodological choices in traditional scientific terms. We cannot fully overcome this challenge.
Part of this section adopts a meta-analytical approach which navigates through the vast body of literature available. We provide recommendations and best practices for each task involved. Where surveys are available, we cite those and provide our take away to the task at hand. 
A high-level overview of the steps involved is presented in Figure~\ref{fig:flowchart_pipeline}, with a detailed description provided in the following text.

\paragraph{Image Preprocessing:}
In this step, Contrast Limited Adaptive Histogram Equalization (CLAHE) is applied~\cite{clahe}. CLAHE is a technique used to enhance the local contrast of an image by dividing it into small tiles and applying histogram equalization to each region. 
This method not only enhances contrast but also ensures more similar contrast across different regions of the same image and across images captured under varying lighting conditions. As a result, it enables the use of a consistent set of parameters for subsequent processing steps, regardless of the input image's lighting variations.

\vspace{-10pt}

\paragraph{Edge Detection:} 
%
%
%
%

Our method is based on the Canny edge algorithm~\cite{canny-algo} and tailored for straight edge-segment detection. The algorithm is applied twice: once for horizontal and once for vertical edges. This separation prevents undesired connections between edges of different orientations, simplifying subsequent steps.
Edge detection begins with applying a Gaussian filter to smooth the input image, followed by gradient computation using Sobel kernels in the horizontal ($G_x$) and vertical ($G_y$) directions. The gradient magnitude and orientation are computed as: \begin{equation*} G = \sqrt{G_x^2 + G_y^2}, \quad \theta = \arctan\left(\frac{G_y}{G_x}\right). \end{equation*}
Detected edges are refined through non-maximal suppression to thin the edges and eliminate spurious responses. Two thresholds ($T_{high}$ and $T_{low}$) are applied during edge tracking to classify pixels as strong or weak edges. Weak edges are retained only if connected to strong edges, ensuring continuity. For our application, thresholds are set to $T_{low}=40$ and $T_{high}=80$.

\begin{equation*}
\begin{aligned}
& \text{(1) Gradient Magnitude and Orientation Check:} \\
& \qquad B(x, y) \equiv \left( G(x, y) > \max(G(x_{\text{nb1}}, y_{\text{nb1}}), G(x_{\text{nb2}}, y_{\text{nb2}})) \right. \\
& \qquad \qquad \qquad \: \left. \text{and} \ \theta(x, y) \in [\theta_{\min}, \theta_{\max}] \right) \\
& \text{(2) Non-maximal Suppression:} \\
& \qquad G(x, y) = \begin{cases} 
G(x, y), & \text{if } B(x, y) \text{,} \\
0, & \text{otherwise.} \end{cases} \\
& \text{(3) Edge Tracking and Thresholding:} \\
& \qquad \text{Edge}(x, y) = \begin{cases} 
\text{Strong}, & \text{if } G(x, y) > T_{\text{high}}, \\
\text{Weak}, & \text{if } T_{\text{low}} \leq G(x, y) \leq T_{\text{high}}, \\
\text{None}, & \text{otherwise.}
\end{cases}
\end{aligned}
\end{equation*}

\vspace{-10pt}

\paragraph{Edge Chaining:}
An approach based on~\cite{opencv-findContours} is used to group edge pixels to continuous edge segments.
This approach performs a raster scan on the input binary image. When an edge pixel is encountered, it follows connected edge pixels until it revisits the initial edge pixel of the segment; thereby efficiently grouping connected edge pixels to edge-segments. Every continuously connected set of edge pixels becomes an edge-segment.
An efficient implementation can be found in the OpenCV library~\cite{opencv_library}

\vspace{-10pt}
\paragraph{Subpixel refinement} 
To extract meaningful results even from short edge-segments, subpixel accuracy is necessary. We recommend the work of Grompone et al.~\cite{2017-subpixel} which is straight forward to implement and improves the work of Devernay~\cite{1995-subpixel-devernay} which occasionally suffers from oscillating artifacts.

\vspace{-10pt}

\begin{figure}
    \centering
    \includegraphics[width=\linewidth]{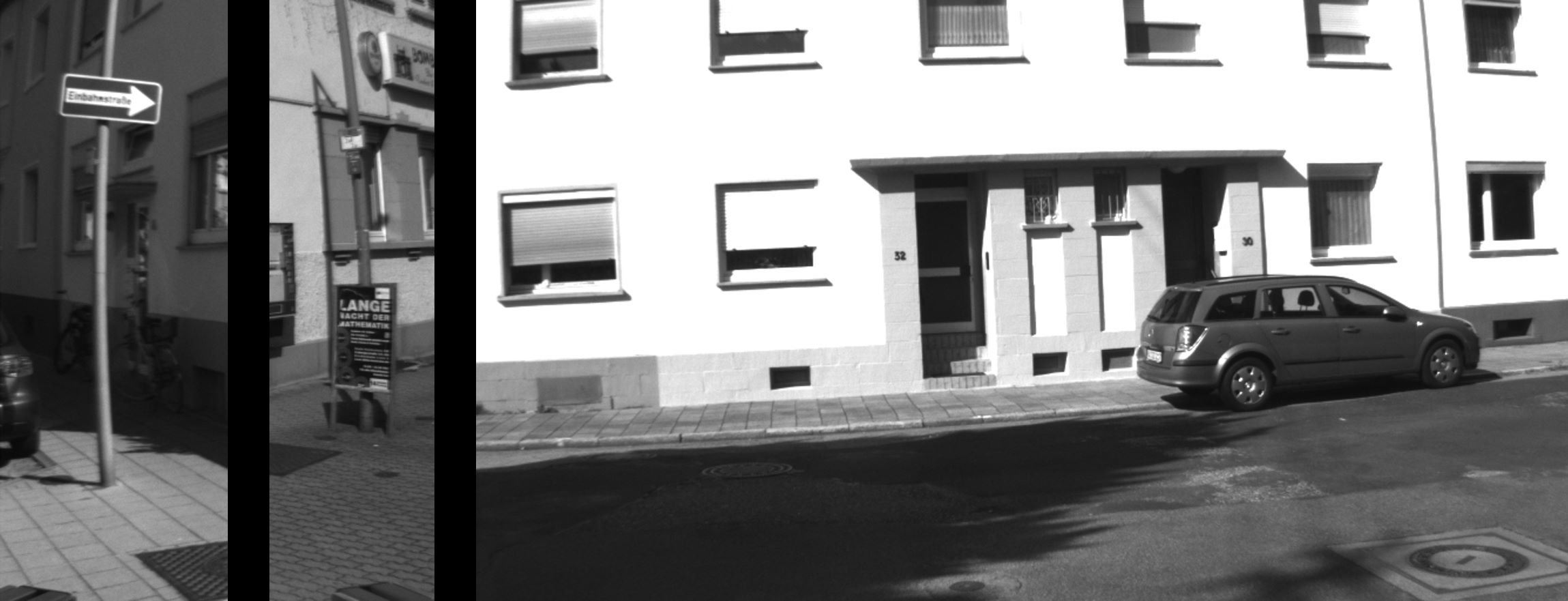}
    \caption{Illustration of 3D straight lines appearing interrupted in the image due to occlusion and discontinuous features. Examples from the Kitti dataset~\cite{2013-kitti-dataset}}
    \label{fig:examples_line_interruption}
\end{figure}

\paragraph{Merging:}
3D straight lines frequently appear discontinuous in images for various reasons. For instance, edge-segments of a series of identically mounted windows 
or poles with additional signs attached to them (see Figure~\ref{fig:examples_line_interruption}). 
To address this issue, we identify and merge edge-segments that are associated with the same 3D line. 
%
We fit a circle to each detected edge-segment, which presents a challenge as the edge-segments often represent only a small fraction of the associated virtual entity. In other words, the extent of the edge-segments is typically just a portion of the radius of the fitted circle.
We employ Taubin's approach~\cite{taubin-circle-fit}, which demonstrates exceptional performance and robustness, converging even in cases of small, nearly straight segments. 
For a comprehensive study on circle fitting techniques, we refer the reader to Al-Sharadqah and Chernov.~\cite{2009-survey-circle-fit}.
%
%
%
For each fitted circle, we calculate the residual to neighboring edge-segments. If it is below one pixel, the edge-segments are associated with the same circle and merged.

\vspace{-10pt}
\paragraph{Filtering:}
The filtering process is divided into the following stages:
i) Shape-Based Filtering:
Edge-segments are first filtered based on their shape. We impose a minimum length requirement of 100 pixels for an edge-segment to be accepted. 
%
ii) Distortion Parameter Estimation:
Robust distortion parameter estimation is then performed to identify edge-segments that conform to a linear approximation after undistortion. Edge-segments that deviate significantly from this approximation are discarded.
To account for distortion, we use the widely recognized Brown-Conrady model~\cite{brown-conrady}, which incorporates both radial and tangential distortion effects. 
The undistorted image coordinates (u,v) are computed as follows:

\vspace{-10pt}

\begin{equation*}
\begin{split}
u = & (u_d-x_c)(k_1r^2+k_2r^4+k_3r^6) + \\ & 2p_1(u_d-x_c)(v_d-y_c) + p_2(r^2+2(u_d-x_c)^2) \\
v = & (v_d-y_c)(k_1r^2+k_2r^4+k_3r^6) + \\ & p_1(r^2+2(v_d-y_c)^2) + 2p_2(u_d-x_c)(v_d - y_c) \\
\end{split}
\end{equation*}

where $r^2 = u^2 + v^2$, $k_1$, $k_2$, and $k_3$ are radial distortion coefficients, $p_1$ and $p_2$ are tangential distortion coefficients and $x_c$ and $y_c$ is the center of distortion. 
To handle outliers effectively, we employ a RANSAC-based scheme~\cite{ransac}. 
The outlier threshold, which defines the distinction between inliers and outliers, plays a critical role in this process. However, guidance on setting this parameter is scattered and ambiguous in the existing literature.
We advocate for the use of MSAC~\cite{2000-msac-mlesac}, a variant that necessitates minimal adjustments to the traditional RANSAC framework. This approach offers a more stable and accurate model ranking, largely unaffected by the specific choice of the outlier-threshold, and consistently yields more precise models.


\section{Discussion \& Future Work} 
\label{sec:discussion}


While our "\datasetname" dataset represents a first step, it currently consists of roughly 150 labeled images. This is sufficient for evaluating algorithm performance but inadequate for training deep learning-based approaches. 
One promising direction for future work is the expansion of the dataset, particularly through the inclusion of synthetic data. Synthetic data offers the advantage of scalability, allowing for the generation of diverse and extensive datasets. However, it also introduces the issue of assuming a lens distortion model. Recent research (e.g.,~\cite{schops2020}) highlights significant limitations in current distortion models, particularly in their ability to accurately represent real-world lens distortion. 
%
Another critical area for future work is the evaluation of camera calibration within the context of SLAM (Simultaneous Localization and Mapping). While straight-line-based calibration methods are effective for estimating distortion parameters, they do not provide perspective parameters, such as focal length and principal point. Fortunately, methods for estimating these perspective parameters from undistorted images exist (e.g., Hartley and Zisserman~\cite{hartley2003multiple}). Combining these methods to derive a complete set of calibration parameters and evaluating their effectiveness within a SLAM framework would be a valuable contribution to the field.





\vspace{5pt}
\section{Conclusion} 
\label{sec:conclusion}


%

This study presents the "\datasetname" dataset, tailored for intrinsic camera calibration using straight lines in real-world outdoor scenarios. Despite its compact size, the dataset addresses challenges in cluttered environments by combining data from KITTI~\cite{2013-kitti-dataset}, IAMCV~\cite{2024-JKU-dataset-2}, an edge-segment detection pipeline, and additional manual labeling. This comprehensive approach offers a starting point for intrinsic calibration evaluation.
The provided methodological guidance and evaluation framework enable researchers to adapt and validate edge-segment detection pipelines, bridging the gap between theory and practical applications.
%
The paper also discusses the dataset's limitations and outlines future directions, including potential expansions and improvements, to support further advancements in online camera calibration.


\section*{Acknowledgment}
This work was funded by the Austrian Research Promotion Agency (FFG), PDrive, project number: 12451001.






\bibliographystyle{IEEEtran}



\end{document}